\documentclass[a4paper]{article}
\usepackage[utf8]{inputenc}
\usepackage{array, xcolor, lipsum}
\usepackage[bottom=2cm,top=2.7cm,left=2.7cm, right=2.7cm]{geometry}
\usepackage{enumitem}
\usepackage{wrapfig}
\usepackage{graphicx}
\usepackage{float}
\usepackage{xcolor}
\usepackage[T1]{fontenc}
\usepackage{wrapfig}
\usepackage{gensymb}
\usepackage[hyphens,spaces,obeyspaces]{url}
\usepackage{natbib}
\usepackage{lineno}

\author{
Jakub Fil$^1$\footnote{Corresponding author--jakub.fil@waiys.com, $^1$WAIYS GmbH, $^2$Zurich University of Applied Sciences, $^3$SpiNNcloud Systems GmbH, $^4$TU Dresden, $^5$Applied Brain Research, $^6$NEOM, $^7$The University of Manchester. }
\and Yulia Sandamirskaya$^2$
\and Hector Gonzalez$^3$
\and Loïc Azzalin$^2$
\and Stefan Glüge$^2$
\and Lukas Friedenstab$^1$
\and Friedrich Wolf$^1$
\and Tim Rosmeisl$^3$
\and Matthias Lohrmann$^3$
\and Mahmoud Akl$^3$
\and Khaleel Khan$^4$
\and Leonie Wolf$^1$
\and Kristin Richter$^1$
\and Holm Puder$^1$
\and Mazhar Ali Bari$^1$
\and Xuan Choo$^5$
\and Noha Alharthi$^6$
\and Michael Hopkins$^1$
\and Mansoor Hanif
\and Christian Mayr$^4$
\and Jens Struckmeier$^1$
\and Steve Furber$^7$
}

\title{Heterogeneous computing platform for real-time robotics.}

\begin{document}
\vspace{-2.5pt}
\maketitle 

\begin{abstract}

After Industry 4.0 has embraced tight integration between machinery (OT), software (IT), and the Internet, creating a web of sensors, data, and algorithms in service of efficient and reliable production, a new concept of Society 5.0 is emerging, in which infrastructure of a city will be instrumented to increase reliability, efficiency, and safety. Robotics will play a pivotal role in enabling this vision that is pioneered by the NEOM initiative - a smart city, co-inhabited by humans and robots. In this paper we explore the computing platform that will be required to enable this vision. We show how we can combine neuromorphic computing hardware, exemplified by the Loihi2 processor used in conjunction with event-based cameras, for sensing and real-time perception and interaction with a local AI compute cluster (GPUs) for high-level language processing, cognition, and task planning. We demonstrate the use of this hybrid computing architecture in an interactive task, in which a humanoid robot plays a musical instrument with a human. Central to our design is the efficient and seamless integration of disparate components, ensuring that the synergy between software and hardware maximizes overall performance and responsiveness. Our proposed system architecture underscores the potential of heterogeneous computing architectures in advancing robotic autonomy and interactive intelligence, pointing toward a future where such integrated systems become the norm in complex, real-time applications.  

\end{abstract}

\section{Introduction}

In the near future cognitive cities will become a reality. They will be meticulously designed for sustainability, powered by renewable energy, and characterized by a digital landscape interwoven with robots coexisting alongside humans \citep{smartcities3040056}. This vision of seamless human-robot interaction will require a one-of-a-kind solution to analyse vast amounts of data powered by energy-efficient and low-latency computing power. 
This project aims to carry out an investigation into heterogeneous combination of brain-inspired hardware with high-density GPU (Graphics processing units) solutions for real-time social robotics. In this paper we propose joining the two brain-inspired approaches: SpiNNaker2 \citep{gonzalez2024spinnaker2} for large-scale brain modeling and Intel's research chip Loihi2 \citep{orchard2021efficient} for real-time perception and actuation at the edge.
Moreover, we combined a Dynamic Vision Sensor (DVS) camera \citep{dvs} with Intel's Loihi2 chip and trained an algorithm to track human hands—marking the first such real-time robotic vision workload deployed on this system. 
We propose to further extend the capabilities of these neuromorphic approaches, by including NVIDIA DGX GPU systems that are used to facilitate showcase orchestration, running large language models and other deep learning algorithms, as well as to execution of a specialized version of the model Spaun \citep{eliasmith2013build, Spaun2_2018}, which uses brain-inspired computation mechanisms to implement a cognitive architecture.
The combination of those elements, humanoid robot, and a functional brain model Spaun, enables performing real-time processing of sensor streams to achieve complex memory tasks.
This multifaceted approach facilitates real-time speech and high-level action prediction, which are critical for the control of a humanoid robot tasked with playing a musical instrument, in an effort to create an interactive demonstration towards an intelligent General Purpose Robot (GPR).
The project involves the implementation of an interactive demonstrator focusing on natural human-robot communication as the first functional capability of an AI enabled humanoid GPR. 
To achieve the outlined goals, we have also designed a highly specialised, 
sustainable supercomputing Micro Data Center with an innovative direct hot liquid cooled cooling system, enabling brain-like capabilities for real-time autonomous systems within a customized environment.

Implementing a real-time AI cognitive platform is crucial for managing vehicles, drones, robots, and other cognitive solutions within future cognitive cities.
In addition to processing vast amounts of data, these platforms must support seamless communication among diverse systems and devices. 
The operation of such a cognitive infrastructure demands advanced algorithms capable of adapting to dynamic urban environments, enabling predictive maintenance, swift anomaly detection, and efficient resource allocation \citep{Shenavarmasouleh2022}. 
It has long been argued that such solution requires a hybrid architecture that combines large-scale deep learning with symbolic reasoning and structured knowledge, supporting the claim that diverse and heterogeneous systems are needed for effective AI \citep{Garcez_neurosym, Marcus2020}. In this paper, we present a collaborative approach that integrates Neuromorphic, Deep Learning, and Symbolic AI, creating a real-time, low-latency, and energy-efficient cognitive AI platform. 
In an effort to exemplify how these elements can be put into practice, we present an innovative use case demonstrating real-time interaction between the robot and a human performing music sequences on a theremin.

We believe that the future of cognitive cities and robotics relies on the integration of heterogeneous computing systems to meet diverse processing needs. High-density GPUs handle compute-intensive tasks over longer timescales, while energy-efficient neuromorphic hardware at the edge excels in real-time, low-latency decision-making. This combination enables optimized performance while reducing energy consumption through the use of brain-inspired algorithms.
Moreover, we believe that it is necessary to also consider other infrastructure elements such as cooling and heat reuse, to enable sustainable operation of a smart city.  
Our innovative direct hot liquid cooling technology complements these systems, enhancing thermal management and further reducing power usage by allowing for efficient heat recycling \citep{struckmeier2018}.
Together, these technologies form a sustainable, energy-efficient platform for the advanced capabilities required in tomorrow's connected cities and intelligent robotics. 

\section{Results}

\subsection{Showcase overview}
 
\begin{figure}[H]
\centering
\includegraphics[width=0.9\textwidth]{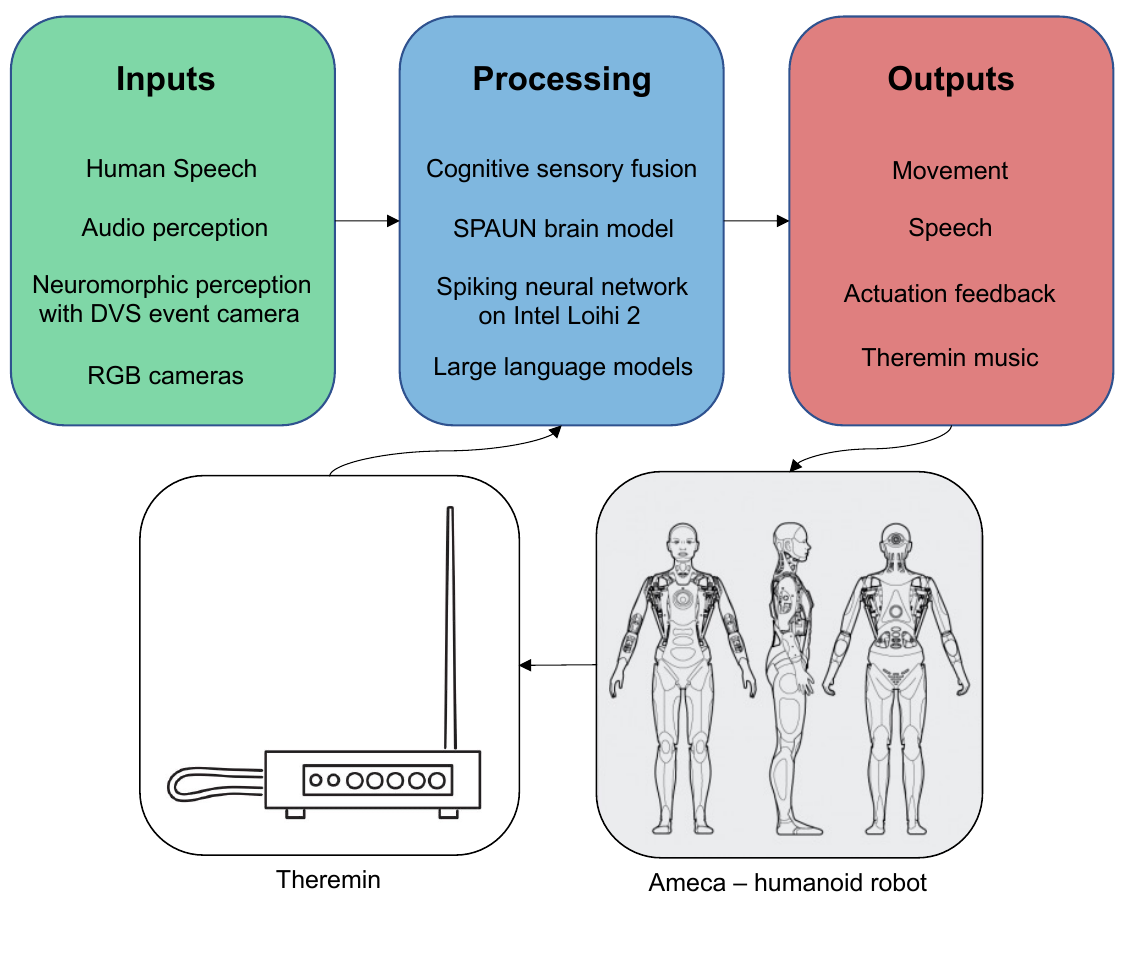}
\caption{Overview of the key components of the interactive musical showcase.}
\label{fig_showcase_software_A}
\end{figure}

Robots are increasingly integrated into our daily environments, and their ability to communicate effectively becomes a crucial consideration. Music, undeniably one of the most potent means of communication, is frequently hailed as a universal language, fostering connections among humans \citep{music_science}. Playing music together requires many skills similar to the ones used in real-world, embodied communication: combination of long-term memory and planning with in-the-moment coordination, turn taking, and multimodal signal exchange.  In this context we use playing a musical instrument as an exemplary task. We teach the humanoid robot Ameca to play music pieces using the theremin, an electronic instrument that can be played without physical contact. Through our showcase, we explore capabilities of the Spaun 2.0 brain model in a DGX-based cluster setup involving multiple GPUs. Moreover, we demonstrate real-time neuromorphic computing capabilities on Intel's research chip Loihi2. Importantly, this work constitutes the first demonstration of integration of a real-time workload on neuromorphic platform with cognitive model, running on a computing cluster, thereby laying the groundwork for advanced real-time vision systems that enable robots to continuously perceive and interact with their environment, while also using a large world model or a knowledge base. 

The core of this use case centers around Ameca, a humanoid robot created by Engineered Arts.
This robot represents a state-of-the-art platform for research in human–robot interaction. Its design incorporates a sophisticated facial mechanism, featuring 32 motors dedicated to enabling nuanced and expressive facial mimicry \citep{Ameca_1}. This capability is integral to the robot’s ability to convey a wide range of emotions. 
This consideration is crucial to make interactions with humans feel more natural by providing nonverbal cues similar to those in human communication \citep{Paulus2013}.
For example, \cite{BREAZEAL2003167} notes, that equipping robots with dynamic and expressive facial features is essential not only for conveying clear affective signals but also for mitigating the uncanny valley effect \citep{Mori_uncannyvalley}, thereby fostering more natural human–robot interactions.

Moreover, Ameca is equipped with flexible and dexterous arms that support complex manipulation tasks.
In this study, the robot was set up to replicate musical notes produced on a theremin by a human reference. Conventional note pre-processing techniques, combined with advanced Deep Neural Network (DNN) models, were employed to ensure precise sound identification. Additionally, the integration of the Spaun 2.0 system facilitated cognitive memory tasks influencing the robot's decision-making processes and the state transitions of the system. Spaun 2.0 is the world’s largest functional brain model  spanning up to 6.5 million neurons, which we deployed using an NVIDIA DGX A100 system. A neuromorphic chip Loihi2 was used for fast on-board processing of visual information to enabling moment-to-moment alignment of the robot with the simultaneously playing human. 

The general objective of this project is to propose a brain-inspired hierarchy of processing combining the cognitive and memory capabilities of large-scale  neural networks models with the energy-efficiency and real-time capabilities of neuromorphic hardware, such as Loihi2, at the edge. Such an unprecedented combination of brain-like systems in the cloud and at the edge has the potential to enable many of the sophisticated use cases required in a cognitive city, while also contributing to the overall goal of reducing the energy consumption of  robotics use-cases at scale.

\subsection{Musical automata and music-playing robots}

The AI community and society at large have been interested in music playing robots for a long time. 
Some very early examples include Ismail al-Jazari - a XII century Muslim pioneer of engineering and cybernetics \citep{al_jazari_music}.
He constructed a musical automaton, in the form of a boat with four automatic musicians, which performed pre-programmed musical sequences.  
More recently, there have been attempts to create robots capable of playing instruments, such as piano \citep{pianist_robot},  violin \citep{violoin_robot}, or percussion \citep{percussionist_robot}. In contrast to these automata, real music playing not only includes faithful reproduction of notes and melodies, but also dynamics that aligns with other players or with individual swings of mood of the player or sound production processes of the instrument in solo performances. It is a dynamic, interactive process.

\begin{wrapfigure}{r}{0.51\textwidth}
\centering
\includegraphics[width=0.51\textwidth]{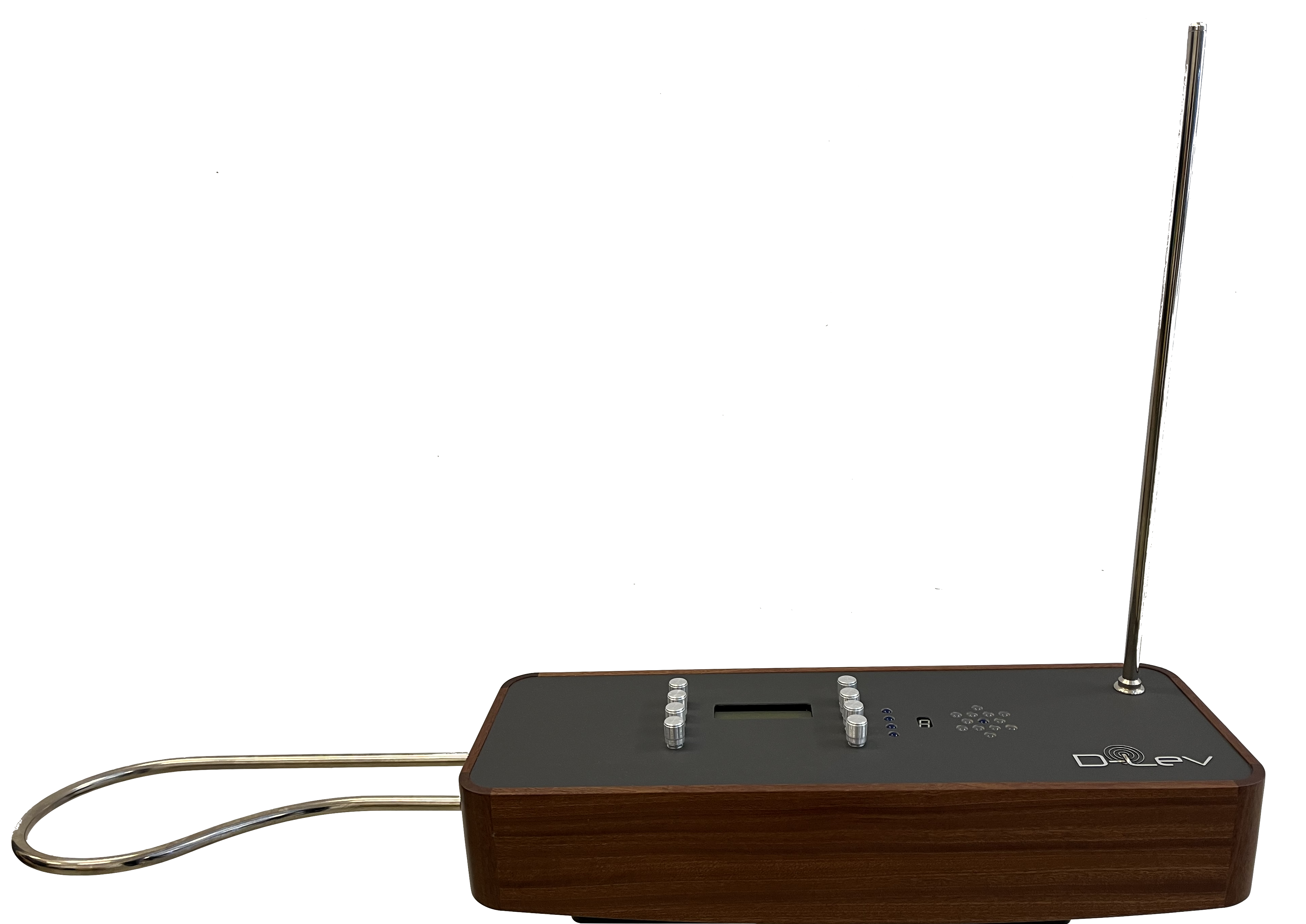}
\caption{The musical instrument used in our study - the Theremin. }
\label{fig_theremin}
\end{wrapfigure}

The theremin, deemed to be one of the most difficult instruments to play, is an electronic musical instrument that requires no physical contact to play music (see Fig. \ref{fig_theremin}).
The theremin produces sound by detecting an interference created  by the player's hands placed between two antennas.
The antenna on the right influences the pitch of the sound, while the one on the left controls the volume. Playing a theremin requires constant adjustment of the hands' positions depending on the produced sound.

There has previously been some research effort devoted to adapting robots to perform music using the theremin, for example iCub \citep{Theremin_2009} and HRP-2 Promet \citep{wu_2010}.
These researchers developed methods for feedforward and feedback arm control based on the theremin’s pitch, which enabled the robots to interact with the theremin in a solo act. The lack of moving fingers allowed only very simple melodies to be played and the robots didn't follow a professionally accepted human-like technique, such as the one coined by Carolina Eyck.

In contrast to the previously described approaches, where robots focused only on pitch fluctuations and ignored the extended dynamic range controlled by the volume antenna, we trained the Ameca robot to control the theremin using its own hands and make decisions using the brain simulator, which we will describe in more detail in the next parts of this paper (see sec. \ref{sec_spaun}). 
These previous approaches also ignored human-like movements such as vibrato, in which subtle frequency changes induce harmonically coupled sounds on top of the base notes. The previous approaches didn't include automatic and advanced calibration, done by the robot itself, which is a must in autonomous systems, especially with an instrument that is inclined to be affected by external actors such as humidity and room geometry. 
Last but not least, none of these robots were equipped with realistic facial expressions to accompany the playing process and increase audience engagement. 
In fact, none of these approaches was designed to be standalone nor to interact with the public. 
Moreover, the approach described in this paper involves cooperative performance between the robot and the user, unlike previous approaches. 
Our system translates the note and volume into movements for both hands.
The backend software provides the robot with the notes to play, as well as their duration, to control the right hand.
Information about the volume is used to control the left hand. 
A sequence of notes and a sequence of volume values are provided to the robot in appropriate states, thus allowing the robot to perform long and complex pieces of music.\footnote{Training process: https://youtube.com/shorts/bj6NaeokJbA} \footnote{Final video demonstration: https://youtu.be/OYW9qQl93aU} 

The robot was trained on a dataset that includes musical pieces played by the master theremin player Carolina Eyck. For each musical piece, the dataset contains video recorded with a ZED2 Stereo camera, data from a Dynamic Vision Sensor (DVS) from iniVation, and the audio. The purpose of this dataset was to generate data that could be used as a reference during the process of translating the melodies into robot movements. 
Images captured via the ZED2 camera were used to perform body tracking.
For this purpose, we used the dynamic pose estimation algorithm from Body Tracking API provided by Stereolabs, creators of the ZED2 camera. 
This system allows for the detection of 18 keypoints on the human body, following COCO18 standard. It allows for accurate prediction of position and movements of human's limbs. Additionally, we used the Object Detection API from Stereolabs to detect the object of interest, and thus further enhance the quality of the recordings. 
Subsequently, we used the resulting body posture recordings to train the robot to perform the same movements, and thus replicate the theremin player. For this purpose, we used a proprietary operating system for Ameca - Tritium, which allows one to program sequences of movements for the robot based on reference recordings.  

In addition to the musical performance, the robot can converse with a person on the stage and with the audience, thanks to speech and language models deployed on the GPU cluster.
Ameca can engage in conversation about playing the theremin, but also about any other topic within the limitations of its language model.
By interacting with the robot, the user can ask for a solo performance, as well as activating teaching or duet mode.
In these modes, the user is presented with the GUI displayed on a tablet in front of the secondary theremin.
The interface shows eight notes ranging from C4 to C5, and the user can engage in an interactive game to accompany the song performed by Ameca.

\section{Discussion}

\subsection{Heterogeneous computing for real-time robotics}


The evolution of social robotics necessitates a computing paradigm that can reconcile the conflicting demands of high-speed signal processing and control and slower long-term cognitive reasoning. In this context, heterogeneous computing platforms—integrating neuromorphic processors with high-density GPU architectures—can be a promising solution. Such systems are essential for enabling seamless human–robot interaction by addressing the diverse computational requirements that span from immediate sensory feedback to complex memory tasks.

A key insight from our work is the importance of aligning software algorithms with hardware architectures. The integration of brain-inspired hardware, such Loihi2, with conventional high-performance GPU clusters demonstrates that a well-matched software–hardware interface is not merely beneficial but essential for efficiency. This alignment minimizes latency in the feedback loop, which is a critical factor when a robot must process sensor inputs, execute decision-making algorithms, and react in real time. As the robot operates over multiple timescales, from rapid perception-action cycles to longer-term planning and learning, the interaction between processing units must ensure that no single component becomes a bottleneck, as demonstrated for example by \cite{ZHAO2019}.
This heterogeneous approach holds significant relevance for both the industry, as well as for the broader neuromorphic and robotics communities. These groups are increasingly attracted to solutions that promise not only technical sophistication but also scalability and energy efficiency. A robust, heterogeneous computing platform addresses market demands by reducing power consumption and operational costs, while enhancing performance in dynamic, real-world environments. Moreover, community-driven efforts in standardizing hardware interfaces and software protocols are laying the groundwork for a future in which disparate systems can communicate effectively, further reducing development risks and fostering innovation.

Integral to this vision is the development of universal communication protocols that can harmonize the interactions between diverse devices and computing nodes. The Transport-Independent Protocol for Universal Communications - Address Event Representation (AER), as proposed by \cite{AER}, may serve as inspiration by leveraging event-based data encoding to envisage a low-overhead, energy-efficient standard for future systems, a feature that is particularly significant for urban robotics where fleet management demands real-time coordinated decision making alongside strict energy efficiency. Despite several attempts to connect event-based sensors and computational units, for example also by \cite{Pedersen2024}, no protocol has become a standard yet, even as efforts are made to transition them from academia into engineering. With SNNs and neuromorphic hardware increasingly pervading the commercial space, there is a strong argument for an easily usable standard that bridges event-based devices and more conventional networks and computational resources. AER is likely to be the underlying high-level protocol, given its flexibility and past success, and the speed and reliability of its software and hardware implementation will be central to any heterogeneous system. A robust protocol must fulfill at least two requirements: 1) a raw, asynchronous transport of spikes without timestamps and minimal overhead for very local, high-bandwidth sensor-to-compute connections; 2) a safer, more conservative transport with reference headers, timestamps, and potentially payloads and error-correction for less local connections subject to corruption, delays, and other uncertainties.

Looking ahead, establishing a common representation and universal communication protocol will be vital for the evolution of social robotics, ensuring that as robots become integral components of cognitive cities, diverse systems can “speak the same language” to facilitate seamless collaboration between high-density cloud-based models and edge-based neuromorphic processors, thereby enabling improved system integration and a more resilient, adaptive robotic infrastructure capable of evolving alongside the increasingly complex demands of urban environments.
Therefore, the anticipated standardization of communication protocols represents a critical step towards realizing a future where integrated cognitive systems form the backbone of smart, sustainable cities.



\subsection{Heterogeneous computing as the future of smart cities}


The urban landscapes of tomorrow will be shaped by integrated, heterogeneous computing systems that bridge the gap between high-performance data analysis and real-time, energy-efficient decision-making. As smart cities evolve, they will increasingly rely on hybrid architectures that combine the computational heft of high-density GPU clusters with the rapid, low-latency responsiveness of neuromorphic processors. This dual strategy enables the simultaneous processing of large models and historical datasets for strategic planning and instantaneous sensor data critical for dynamic urban management.

Many scholars have explored the foundational elements of smart cities. For instance, \citep{Caragli} and \citep{Albino} provide comprehensive frameworks that emphasize scalability, resilience, and the transformative potential of integrating advanced technologies into urban infrastructures. Their insights highlight the necessity for systems that not only manage everyday urban functions but also anticipate and adapt to future challenges.

Building on these foundational insights, it is evident that the multifaceted challenges of future urban environments will demand heterogeneous computing architectures capable of spanning a wide range of operational timescales and decision-making processes. For instance, the instantaneous processing of edge sensory data, crucial for real-time traffic management, environmental monitoring, and emergency response, must be integrated with long-term analytics that underpin urban planning. Furthermore, effective autonomous fleet management, including service robots, delivery drones, and self-driving vehicles, relies on systems that can respond quickly to dynamic changes while continuously learning to optimize performance. Nevertheless, conventional deep learning models often struggle with continuous learning and often suffer catastrophic forgetting, as highlighted by \cite{liu2017}. One promising alternative is to employ brain-inspired approaches, such as BitBrain \citep{BitBrain}, which have a natural ability to learn quickly from small amounts of data and then retain, adapt, or forget this knowledge over time as required.
In parallel, fostering effective human-robot collaboration in public safety, healthcare, and municipal services calls for computational frameworks that merge context-aware responses with predictive decision-making capabilities. This diversity of use cases underscores the importance of adopting heterogeneous computing solutions that are agile and capable of adapting to the evolving complexities of smart cities \citep{Zanella2014, Batty2012}.

Furthermore, innovations in thermal management—such as direct hot liquid cooling—are not only optimizing the thermal performance of centralized and distributed computing resources but also paving the way for innovative urban energy recycling. For instance, the excess heat generated by state-of-the-art data centers can be redirected into district heating networks to warm residential buildings, public facilities, or even urban greenhouses. In some pioneering projects, waste heat has been used to preheat domestic water or power efficient cooling for robotics fleets and autonomous vehicles, thereby reducing both energy consumption and operational costs. This synergy between high-performance computing and sustainable energy reuse is a cornerstone in designing smart cities of the future (see sec. \ref{cooling}).

In summary, we believe that the convergence of heterogeneous computing architectures, standardized communication protocols, and advanced energy management solutions offers a robust blueprint for the smart cities of the future. This integrated approach will not only empower urban planners to optimize resource allocation and infrastructure resilience but will also enable intelligent robotics to operate more efficiently, ensuring that tomorrow’s cities are both smart and sustainable.


\subsection{Brain-inspired approaches in combination with machine learning}

For the purpose of enhancing decision-making sub-system of the robot we employed a brain-inspired approach - Spaun 2.0 (Semantic Pointer Architecture: Unified Network). It is a large-scale cognitive neural model consisting of 6.5 million neurons \citep{Spaun2_2018}.
This model was used for cognitive memory tasks and the robot's control process, therefore allowing Ameca to interact with the users using a set of predefined scenarios, i.e. for playing solo, teaching how to play the instrument, conversing etc. 
For the purpose of this project we deployed the Spaun model on a GPU system using Nengo, which is an open-source neural simulation platform that enables researchers to build, simulate, and deploy large-scale cognitive models on both conventional computers and neuromorphic hardware \citep{nengo}.
In the future, we envision that a server with 5 SpiNNaker2 modules could replace the
execution of the Spaun model and further reduce the energy footprint.
See Section \ref{sec_spaun} for more details about our usage and GPU-based implementation of this model.

Although Nengo and Spaun are made up of a complex and ingenious interaction between different modules and computational paradigms inspired by brain connectivity and population coding, much of the representational power comes from a choice of one of the Vector Symbolic Architecture (VSA) types which underlies the operation of Spaun. The breadth of possibilities allowed by VSAs is indeed significantly greater than this one choice and it is perhaps not widely appreciated how flexible these ideas are \citep{Kleyko_2022, Kleyko_2023, Schlegel_2021} or how well they can be applied to actual machine learning and AI problems relevant to social robotics such as visual scene recognition by \cite{Renner_2024} and classification, as described by \cite{Ge2020ClassificationUH}, amongst many other cognitive architectures such as: ACT-R \citep{ritter2019act}, SOAR \citep{laird2019soar}, Adaptive Resonance Theory \citep{grossberg2013adaptive}, or DAC \citep{moulin2017dac}.

As well as possessing desirable levels of robustness and viable solutions to the 'binding problem'--which are two of the key issues for AI systems if they are going to compete with biological intelligence--they are also arguably a very natural fit to event-based sensors and computational resources and so should allow the full latency and energy benefits of neuromorphic computation to be realised.  Although it has not yet been proven by demonstration, a direct, sparse, binary, event-driven implementation of a well-chosen set of VSA mechanisms should be significantly faster and more energy efficient than the Spaun implementation using Nengo with its large populations of rate-coded neurons and associated encoding/decoding computations.
This is because very simple and massively parallel local homeostatic and structural plasticity mechanisms inspired directly by Dendritic Computation principles \citep{subutai2016, BRANCO2010494, KASTELLAKIS201519, LARKUM2008321, London2005DendriticC} have been shown to autonomously generate sparse, binary patterns which are ideally suited to appropriate VSA mechanisms with little or no intervening computation or data movement \citep{BitBrain}. The savings are even more obvious in the context of a more conventional ANN where hundreds or thousands of epochs of backpropagation and the necessary per-epoch floating point calculations of derivatives are extremely time- and energy-intensive




\subsection{Importance of neuromorphic computing}

Most computing systems today are based on the classical von Neumann computer architecture in which a processor, or a central computing unit (CPU), performs elementary computations sequentially fetching variables from memory, one at a time (or at least in dense blocks of contiguous memory). Billions of operations per second can be performed this way, controlled by a central computer clock ticking at multiple Gigahertz. Neural network based algorithms that power today's AI require massively parallel computation, as their computing substrate comprises millions of neurons and billions of connections. When such algorithms are implemented on a CPU, a lot of time and energy are wasted on moving variables out and back to the memory, for every single computation. Modern GPUs alleviate part of this problem by introducing a parallel system of many computing units that still fetch the data sequentially from a separate memory unit. This allows neural network algorithms with appropriate network architecture to run much faster than on a CPU. However, a lot of energy is still wasted on transferring variables to and from the memory, especially if the structure of the neural network can not be mapped easily on the system of local memory (caches) on the GPU. 


Neuromorphic technology is the next fundamental step in computing architecture. Its structure is inspired by the neural networks in biological brains. 
Biological brains have evolved to efficiently combine real-time perception and control with long-term memory, formed and conserved on different time scales.  Human brains require only 20-30 Watts to function, to ``compute" over 86 billion of neurons and trillions of synaptic connections. 
These networks form massively parallel systems of elementary computing units and circuits, each of which ``ticks'' independently of the others i.e., there is no central clock triggering synchronous computation and the whole system supports a more efficient, asynchronous computing mode that can be replicated in hardware. Moreover, as in the brain where computing units are inseparable from the memory -- both are realised in synaptic connection between neurons as well as their internal states -- neuromorphic hardware systems feature local memory, co-located with the processing unit (as in digital neuromorphic devices, such as SpiNNaker or Loihi) or embedded in the computing VLSI circuits of mixed-signal devices or in-memory computing systems.  Thus, neuromorphic processors compute with a fine-grained parallelism that does not rely on a particular neuronal network topology (as GPUs that are well-suited for batched, dense matrix- and tensor-based computation). This property allows for low-latency power-efficient processing \citep{sandamirskaya2022neuromorphic}. 
\cite{Renner_2024} shows that for large network sizes Loihi can be up to 171 times more efficient in terms of energy-delay-product (EDP).
Other researches showed that a EGRU-based language model can be simulated with SpiNNaker2 using only 0.39W, while an NVIDIA A100 GPU would need 60W for an equivalent task \citep{nazeer2024language}.
At the same time, it was shown that the first generation of SpiNNaker can be used for low-latency simulation of different areas of the brain. 
SpiNNaker allowed for the real-time simulation of the sensory cortex, surpassing best-published efforts on HPC by the factor of 3, and on GPUs by a factor of 2 \citep{Rhodes_2019}.

The importance of considering alternative computing paradigms, such as neuromorphic, is especially apparent when we consider energy efficiency. In our project, we chose to use Loihi2, rather than basing the hand detection subsystem on a GPU-based system such as Nvidia Jason Nano. 
Jason Nano has a maximum power consumption of 5-10W, depending on the setup, while Loihi2 only requires 4mW to drive the computation.
The individual SpiNNaker2 chip achieves an energy consumption of the order of 1 to 2.5W and in a range of 48-120W per 48-node board \citep{Rhodes_2019}. For the purpose of this paper, we have estimated that SPAUN implementation on SpiNNaker2 would take ten 48-node boards, as compared to a full DGX with 8 A100 GPUs. The DGX A100 unit consumes about 6.5KW, therefore a SpiNNaker2 implementation would consume between 5.4 to 13.5 times less power.
Another advantage is that on the SpiNNaker platform, the model could be run in real-time, whereas using the DGX box, the model had to be deployed at 45\% of real-time.
Although that was sufficient for our use case, this could prove to be problematic for real-time applications.

Neuromorphic hardware opens a new algorithmic space, in which event-based, or spiking neural networks with various topologies can solve perception, memory formation (i.e. learning), planning, and control tasks orders of magnitude faster and more power efficiently than conventional algorithms. New ways of computation and information representation -- such as temporal computing with spike-events and spike-timing dependent plasticity, or spiking realisation of vector-symbolic architectures and hyperdimensional computing, or attractor dynamics based computation with neural fields -- mark a new era of computing theory and technology. These methods allow powerful computations to be carried out even with small models, enabling rapid recognition and tracking of patterns, dynamic memory formation, and real-time control. 
There are many other ways in which biologically-inspired learning and inference mechanisms will offer new opportunities, such as fast and continuous learning from small amounts of data which because of how current AI models are constructed and learned will always be difficult for them to achieve. The 'binding problem' has also proven to be very difficult to solve in conventional artificial neural networks (ANNs).
Other less obvious benefits that may be of equal importance are the ability to work robustly in the presence of corrupted data or sensor degradation \citep{BitBrain}. These capabilities are particularly useful to enable future ubiquitous robotic systems that will require a lot of intelligent computation, performed in real time in a sustainable manner.  
Despite the many computational and inferential opportunities offered by biologically-inspired, SNN-like and neuromorphic algorithms, the tiny amount of R\&D investment so far spent on them compared to more conventional ANNs and LLMs means that they continue to lag behind in terms of absolute performance in most cases. A particular problem is that of toolchains and frameworks which allow experimentation and sharing of new ideas and which are very well developed in the more conventional AI space. As these new ideas draw in more interest and investment we will undoubtedly see a leveling up of the playing field.
The promising results already achieved will almost certainly be consolidated with new ones that compete much more closely whilst retaining their engineering advantages, as long as those with vested interests in heavily memory-reliant and energy-hungry ANNs do not attempt to stymie this competition.

\section{Methods}

\subsection{Hardware overview}

\subsubsection{Ameca - General purpose social robot}
\label{sec_ameca}
\begin{wrapfigure}{r}{0.5\textwidth}
\includegraphics[width=0.5\textwidth]{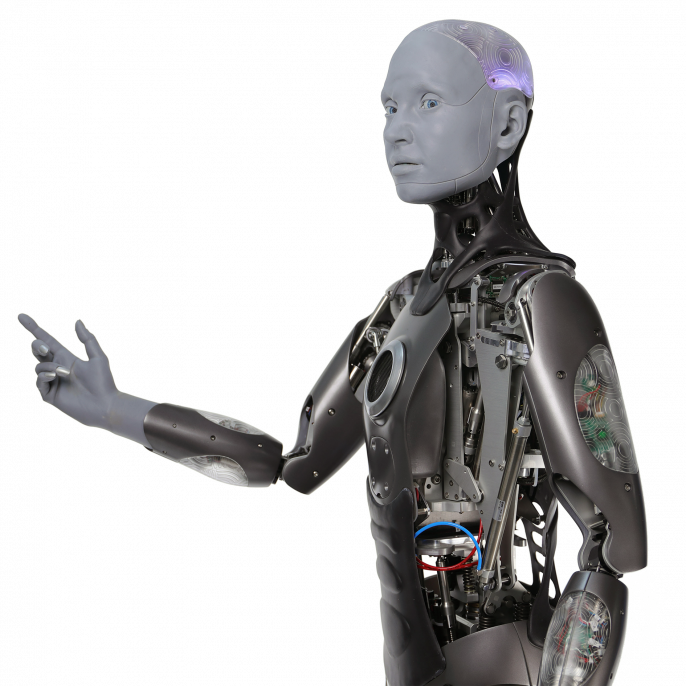}
\caption{Ameca - the humanoid social robot produced by Engineered Arts.}
\label{fig_ameca_figure_A}
\end{wrapfigure}
Ameca is a social humanoid robot which can exhibit life-like human facial expressions \citep{Ameca_1}.
It has been described as the world’s most advanced social robot.
The robot was developed by the robotics company Engineered Arts, the leading designer and manufacturer of humanoid entertainment robots.


Ameca uses a proprietary operating system, Tritium, which allows for extending its capabilities via Python scripts. In order to make decisions on its movement and reactions, it is endowed with a buffering system and connected to the Spaun model responsible for state transitions. 
Ameca collects environmental data through hardware components, such as motors, sensors, cameras, and microphones. The robot can communicate via a speaker built into its chest and an external microphone added for interactions between the user and the robot. The microphone is directly connected to the robot's internal computer - an Intel NUC - to reduce background noise and improve the quality of the transferred audio.

\subsubsection{Micro Data Center}

A Micro Data Center (MDC) provides the backend of the showcase. 
To minimize the acoustic noise in the stage area that is emitted from the server ventilation the MDC was built as a soundproofing housing.   
The MDC consists of two racks including the following components: 

\begin{itemize}
\item	3 NVIDIA DGX A100 servers,
\item	 Data server with Dual CPU AMD Epyc 7313,
\item	 Video server for displaying visual content,
\item	 Network switches for management access to the backend and network integration of frontend devices and Intelligent Platform Management Interface (IPMI) for the server.
\item Peripheral infrastructure including: electrical and control infrastructure, fire suppression and ventilation systems, amplifier and mixer of the audio system.
\end{itemize}

\subsubsection{GPUs}
\label{dgx}


Three NVIDIA DGX A100 systems, each equipped with 8 A100 GPUs were used in this project. Each system was dedicated to a specific functionality:

\begin{itemize}
\item DGX 1: Large language model and speech recognition.

The first DGX system hosts the 
large language model (LLM) with 13 billion parameters
. It is used for low-latency general knowledge conversation with the robot. The input for the LLM is provided by the speech recognition engine - Whisper by OpenAI. 

\item DGX 2: Spaun 2.0 model and audio processing.

The DGX 2 is used to host the Spaun 2.0 model.
which was adapted to run on the A100 GPU architecture using 7 GPUs (see section \ref{sec_spaun}). 
The audio input from the two theremins is provided to DGX-2 using USB interface.

\item DGX 3: Processing sensory input for stage monitoring and lights management.  

The third DGX handles the detection 
of people and their position relative to the stage.
The sensor input is provided by 3 cameras (2 RGB cameras, 1 depth camera). Each camera is used for a designated task (see sec. \ref{cameras}).

\end{itemize}
\subsubsection{Neuromorphic}

\paragraph{SpiNNaker}
The SpiNNaker system \citep{SpiNNakerBook} is a digital neuromorphic platform developed to simulate large-scale spiking neural networks. The installation built at the University of Manchester hosts over 1 million ARM968 cores in a single cluster, thus allowing for an unmatched parallelism, vastly improving the speed of computation compared to standard processor architectures. This made the SpiNNaker neuromorphic cluster in Manchester the world’s largest real-time brain simulator, until it was recently overtaken by the Hala Point system based on Loihi2.
However, the new SpiNNaker2 cluster being built by the Technical University of Dresden exceeds the capacity of those two systems, setting up a new record for the world's largest brain-like supercomputer. 


\paragraph{SpiNNaker2}
To unleash the cutting-edge real-time autonomous capabilities, in the next iterations of the project we will harness the power of SpiNNaker2, the most flexible neural architecture in the realm of neuromorphic substrates. SpiNNaker2 serves as an accelerator tailored for extensive event-based and asynchronous processing \citep{mayr2019spinnaker}. 
Thus, it allows for unprecedented energy-consumption reduction, reaching 1/10th of the power used in a comparable GPU setup.

The SpiNNaker2 microchip comprises 152 ARM-based processing elements interconnected via a network-on-chip (NoC).
The chip's collective 19 MB on-chip SRAM is complemented by 2 GB of LPDDR4 memory. 
What sets it apart from the first generation SpiNNaker chip is specialized accelerators for exponential functions, random number generation, and multiply-accumulate (MAC) operations. 
These features allow it to support the efficient implementation of sparsity-aware artificial neural networks, symbolic AI, and spiking neural networks \citep{gonzalez2024spinnaker2}. 
SpiNNaker2‘s biological inspiration is evident in the chip’s architecture. Distributed processing elements work asynchronously, enabling massively-parallel event-based and sparse computation with on-demand power consumption. 
The chip is designed to minimise power consumption during idle times compared to peak performance. Each processing element is running autonomously, allowing power control with Dynamic Voltage and Frequency Scaling (DVFS) per element, which ensures fine-grained power optimisation on-demand.
A light-weight Network-on-Chip distributes information across the chip and supports input and output streaming of data with a minimum latency.
Brain-like accelerators enhance computational performance for biologically inspired networks. Hence, the system provides native support for hybrids of traditional AI and spiking networks.

\paragraph{Sustainable Supercomputing capabilities and hot liquid-cooling design}
\label{cooling}

The expected exponential growth in AI application and hardware in the next few years makes it very meaningful to investigate energy efficient solutions to save $\textrm{CO}_2$ emissions and operational costs. 
One pillar of this approach is the use of neuromorphic hardware. Another is the implementation of hot liquid cooling that enables a maximum efficient operation of the cooling infrastructure and an optional waste heat reuse. 
A reference example is the design of the 100\% liquid-cooled infrastructure for SpiNNaker2 in cooperation with TU Dresden and SpiNNcloud systems.
The patented liquid cooling technology designed by Cloud\&Heat and WAIYS not only operates much more quietly than conventional air cooling but also leverages high liquid temperatures to enable effective waste heat reuse for facility cooling or even industrial process heat applications \citep{struckmeier2018}.
Our previous datacenter development has shown that the cooling circuit for a neurmorphic server consumes approximately 1\% of the total energy needed for the whole cluster, which significantly outperforms any air-cooled solutions.
When this technology is combined with other complementary cooling methods, the integrated approach can reduce energy consumption by up to 70\% compared to traditional air cooling, underscoring the importance of employing a diverse cooling strategy for energy efficient use cases \citep{cloudheat2019}.

\paragraph{Intel Loihi2 - Real-time Processing of Sensory Streams}
\label{realtime_proc}

The Loihi chip was originally released by Intel in 2018, along with the launch of Intel's Neuromorphic Research Community program
\citep{davies_intel}.  Loihi2 was released in 2021 alongside Lava - an open-source software framework for developing neuro-inspired applications \citep{orchard2021efficient}. Loihi2 is a neuromorphic chip that generalizes the spiking neural network computation to asynchronous graphs with almost arbitrary topology at scale and adds the capacity to send graded information as spikes. This greatly expands the solution space available to algorithm designs. Lava opens a new way to think of neuromorphic algorithms as a system of interacting processes, each unfolding in time and exchanging spike-messages with other processes running on neuromorphic or conventional hardware, or with the interface to the physical world (sensors, robot controllers, or screens).   

The focus of our use of Intel technology in this paper is to demonstrate how this emerging computing hardware can benefit robotic applications. Owing to its fundamentally different way of performing computations in an event-based, asynchronous manner, its fine-grained parallelism, and on-chip learning, neuromorphic hardware enables a new kind of robotic functionality: namely to integrate sophisticated ``intelligent" processing with artificial neural networks -- as required for object detection, tracking, and localization, as well as motion planning -- in fast real-time control loops. This fast perception and cognition capability will make robots safer and more effective in dynamic, human-centered environments. In the showcase, we focus on the sensing part of this processing, demonstrating fast, real-time visual processing. 

A crucial component of enabling real-time vision in our show case is the neuromorphic, event-based camera - Dynamic Vision Sensor (DVS). A DVS pairs particularly well with neuromorphic computing hardware: both events of the DVS and spikes in neuromorphic chips are asynchronous signals, making visual computation energy efficient and fast. We have integrated a DVS sensor with Intel's Loihi2 chip and have trained a neural-network algorithm to track human hands based on asynchronous change events, output by the DVS. This is the first time that such real-time workload is demonstrated on Loihi, paving the way towards real-time vision, which is required for robots to be aware of their environment on a moment-to-moment basis. 


\begin{figure}[H]
\centering
\includegraphics[width=0.9\textwidth]{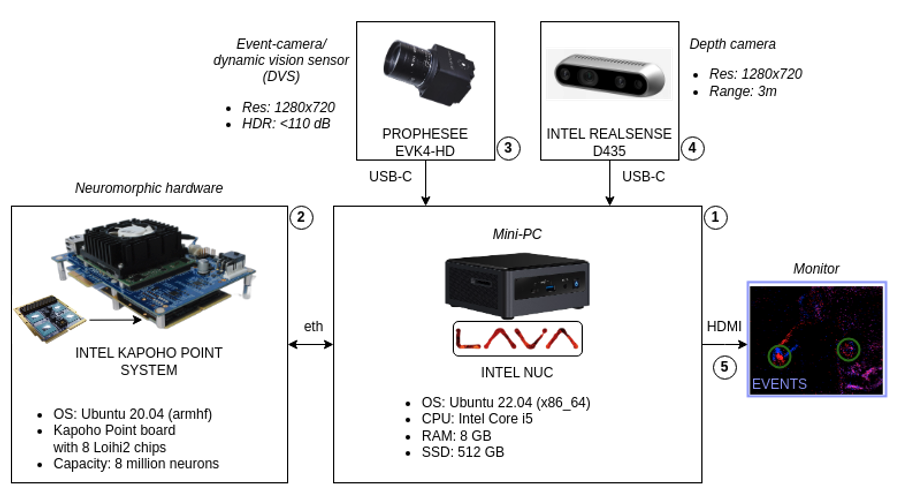}
\caption{Overview of the hardware layer for the real-time neuromorphic hand tracking.}
\label{fig_loihi_hardware}
\end{figure}

The hand tracking system was constructed using off-the-shelf components, in combination with the most recent Intel's development board, Kapoho Point, and neuromorphic research chip Loihi2, see fig. \ref{fig_loihi_hardware}. The delivered hardware solution consists of a number of key components:

\begin{itemize}
\item The Kapoho Point system - Intel’s neuromorphic development board, consisting of 8 Loihi2 chips, an Intel FPGA system, and a baseboard to handle peripherals. One of the peripherals is an ethernet interface allowing communication with the NUC. The ethernet interface is used to send the commands to the neuromorphic chips from the NUC as well as to relay any data generated by the on-board application.

\item Prophesee Evaluation Kit 4 HD (EVK4-HD) is an event-based camera which outputs spatiotemporal streams of events with 1280x720 spatial resolution and with a latency of 220$\mu$s with a high dynamic range greater than 110dB. 

\item Depth information is captured by the Intel RealSense Depth Camera D435 at a resolution of 1280x720 and up to a range of 3m.

\end{itemize}


\subsection{Software overview}

In this section, we will dive deeper into the software and hardware architecture which enables the unique interactive musical experience, as well as discuss some key components which allow Ameca to communicate in a seamless way. 
The robot is provided with a variety of environmental cues, coming from the sensors placed directly in the robot, and additional sensors placed around the stage.
Some of these additional sensory inputs come from the cameras placed at different positions around the stage, others from additional microphones on the stage and next to it. 
Additionally, the robot receives the digital information from the theremins, in order to control its sound and aid the user in the interactive duet mode.

\begin{figure}[H]
\centering
\includegraphics[width=1\textwidth]{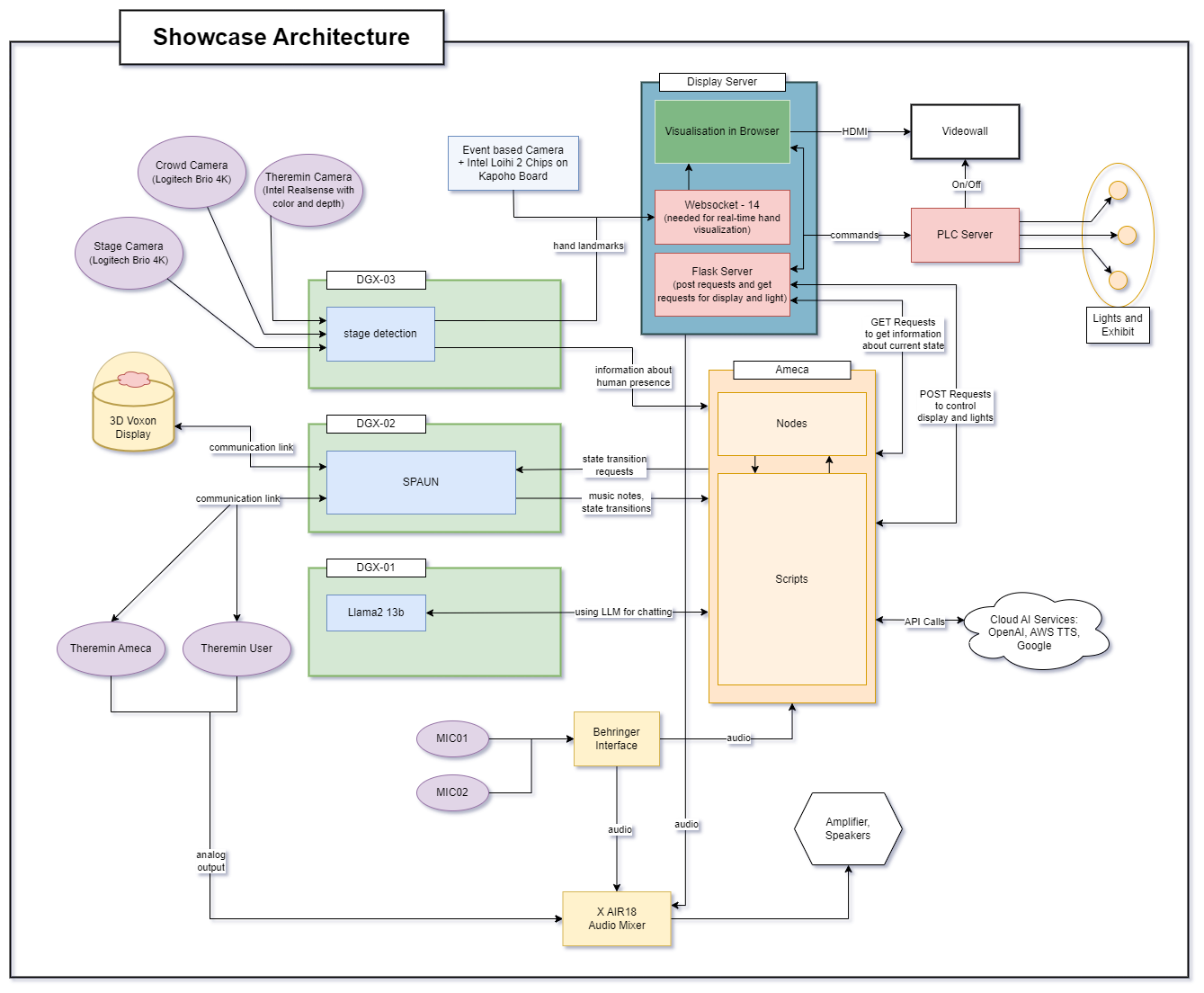}
\caption{Overview of the hardware and software integration architecture for the interactive musical showcase.  
}
\label{fig_showcase_architecture}
\end{figure}

Figure \ref{fig_showcase_architecture} depicts all of the hardware and software components built for the purpose of this project, and how are they integrated.
We employed a unique integration of state-of-the-art systems to process the incoming streams of data. This combination involves neuromorphic solutions, such as Loihi2 for low-latency hand detection, high-performance GPUs in the form of NVIDIA DGX A100, which enables local simulation of the large language and speech recognition models, as well as cloud solutions.
In addition to that, the architecture includes the stage management system, which allows for the control of the stage lighting and video wall display.
Moreover, we have integrated a number of audio processing elements, including speakers, microphones, and interfaces.

\subsubsection{Spaun 2.0 integration for cognitive memory tasks and role in decision-making}
\label{sec_spaun}

To emulate the state transitions of the system a Spaun 2.0 model was used.
Spaun (Semantic Pointer Architecture: Unified Network) is a large-scale cognitive neural model consisting of 2.5 million neurons \citep{Spaun_2012}, implemented using the Semantic Pointer Architecture \citep{eliasmith2013build}.
This architecture was later expanded in Spaun 2.0, the World’s Largest Functional Brain Model \citep{Spaun2_2018}.
This unique model has up to 6.5 million neurons used for cognitive memory tasks and to further enhance the robot's control process, thus allowing Ameca to interact with the users using a finite-state machine (FSM).
The model can perform 12 cognitive tasks and has been demonstrated to reproduce behavioural and neurological data observed in natural cognitive agents.
The functional modules of the model corresponded to specific regions of the brain. 
The authors demonstrated the feasibility of establishing models to conduct various tasks by following the organisational principles of the human brain.
The Spaun model is comprised of 8 modules: visual processing, information encoding, transformation, reward evaluation, information decoding, motor control, working memory, and action selection.
Spaun’s action selection module is responsible for keeping track of Spaun’s progress and deciding what actions to perform as it executes each task. Additionally, the action selection system is responsible for managing the flow of information throughout the different networks that make up Spaun.

\begin{figure}[H]
\centering
\includegraphics[width=0.9\textwidth]{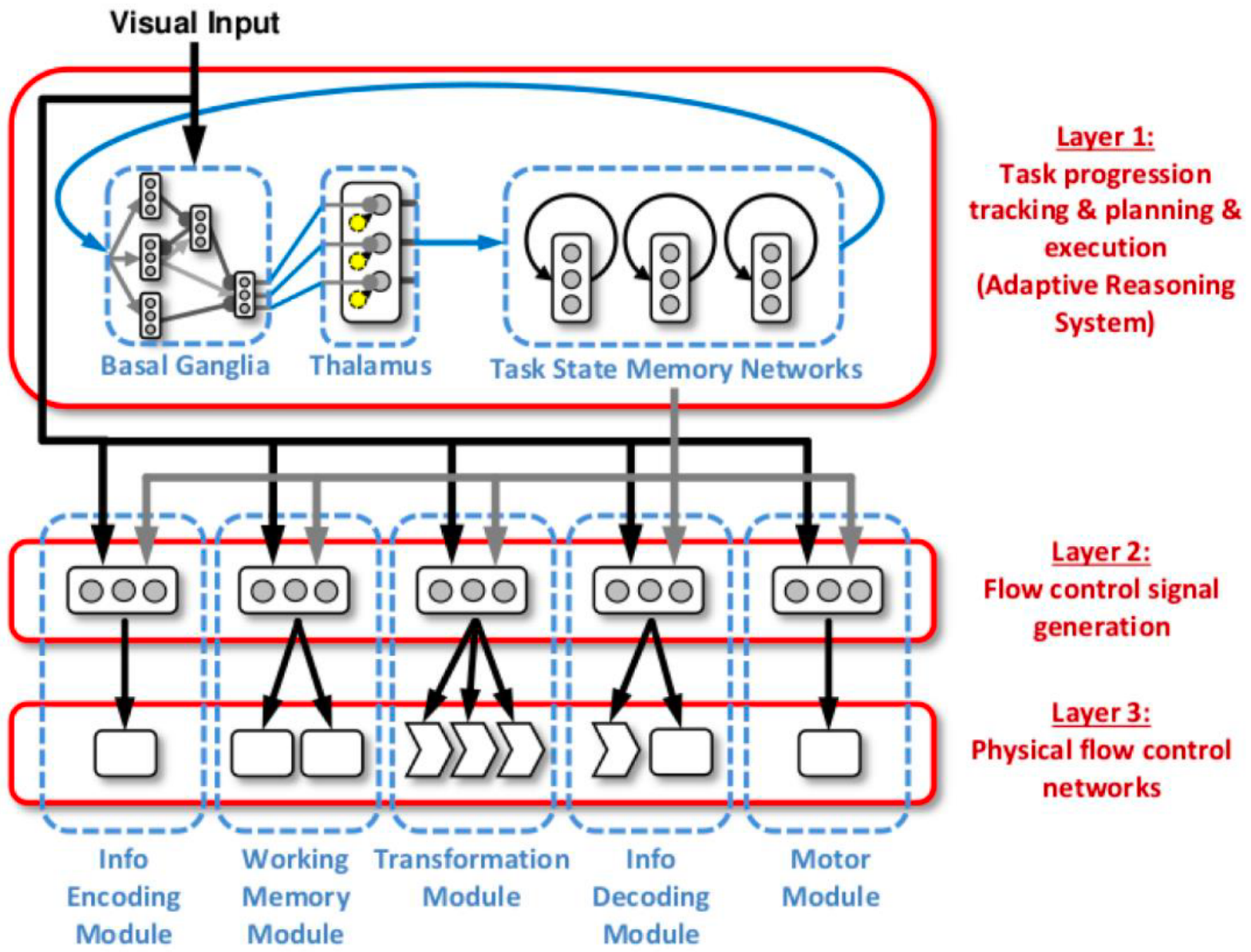}
\caption{Schematic diagram of the three-layer action selection hierarchy (Diagram from \cite{Spaun2_2018}).}
\label{fig_spaun_action}
\end{figure}

\begin{figure}[H]
\centering
\includegraphics[width=0.9\textwidth]{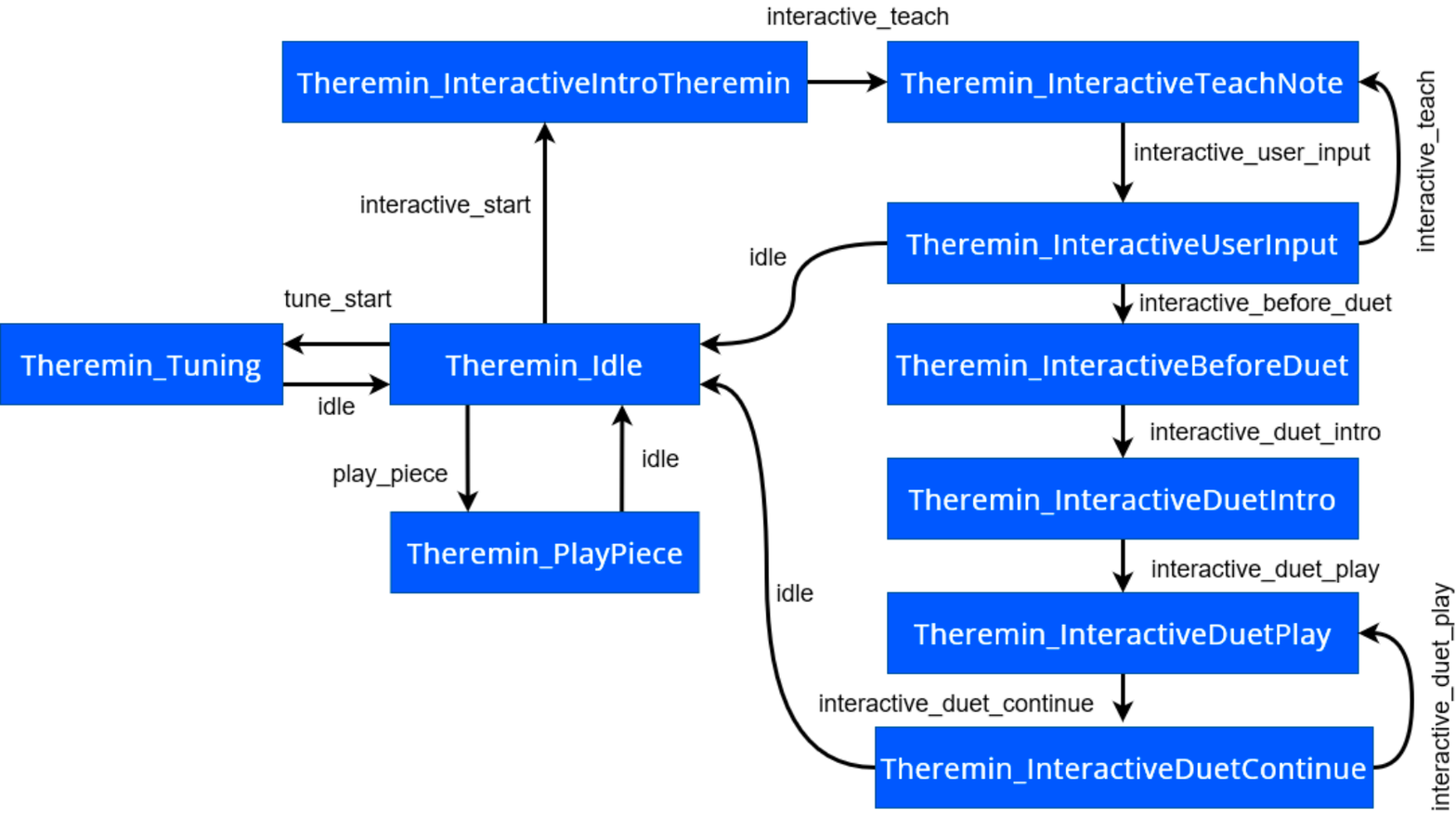}
\caption{Overview of the robot's state transitions relevant for the theremin interactions.}
\label{fig_theremin_states}
\end{figure}

In order to reduce the number and complexity of the condition-consequence pairs, Spaun’s action selection module is implemented as a three-layer hierarchy, with each layer of the hierarchy responsible for different functions of the action selection system (see fig. \ref{fig_spaun_action}). The top level of the hierarchy receives input from the vision module and is only responsible for high-level tracking, planning, and execution of Spaun’s task progress. This layer is implemented using the reasoning system. The second layer of the hierarchy combines the state representation of the top layer with the visual semantic pointers to generate the module-specific control signals used to control the flow of information within each of Spaun’s modules. The last layer of the action selection hierarchy includes the physical networks necessary to accomplish the information flows dictated by the outputs of the second layer of the action selection hierarchy. 
The behaviour of the state machine, and thus the decision-making of the robot, is controlled by the Spaun model.  
This allows the robot to modify its behaviour according to a system of state-based tasks related to music playing (see fig. \ref{fig_theremin_states}).

\begin{figure}[H]
\centering
\includegraphics[width=1\textwidth]{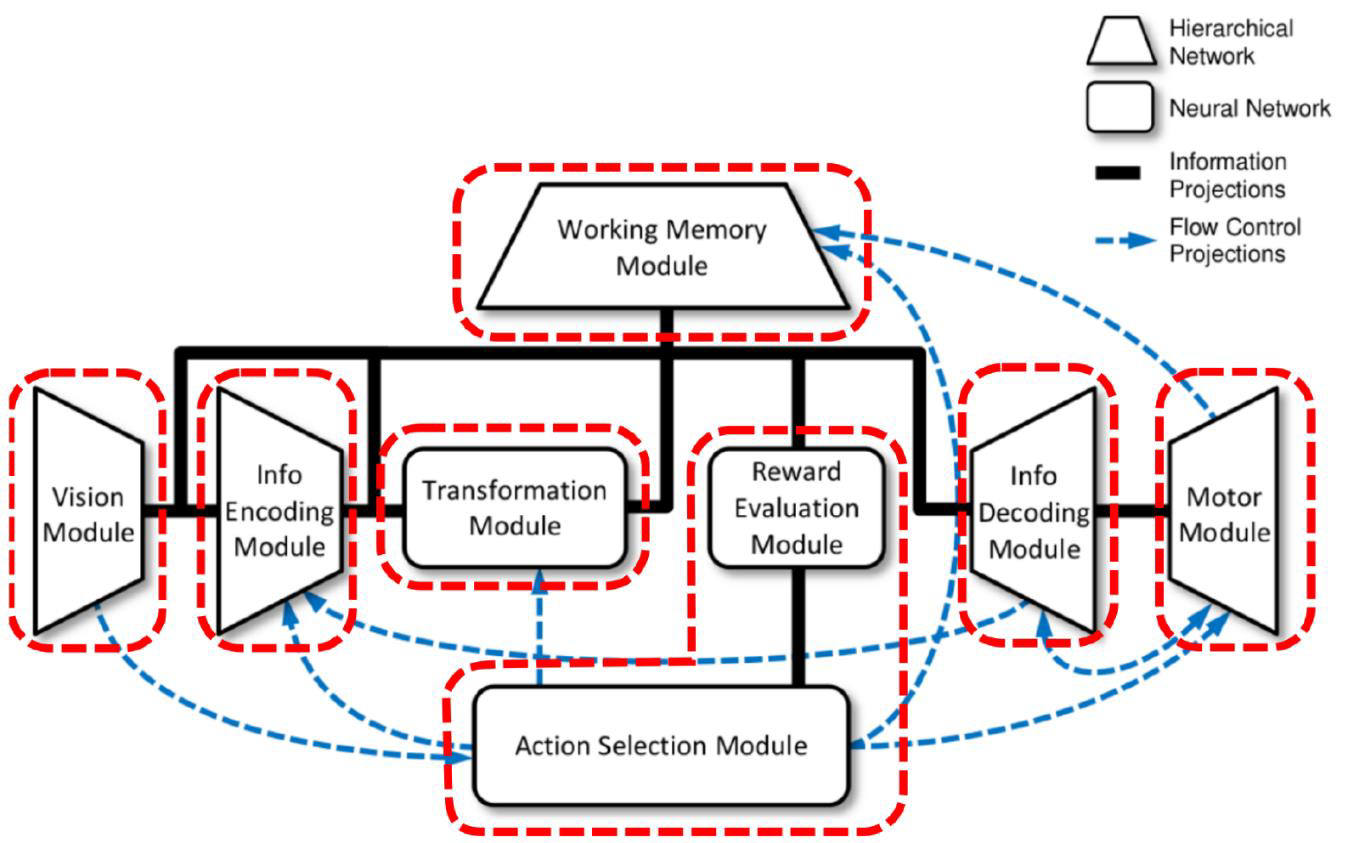}
\caption{The segmentation of the Spaun 2.0 model architecture across 7 GPUs. Each red box indicates a part of the model running on its own GPU (Diagram from \cite{Spaun2_2018}).}
\label{fig_spaun_gpu}
\end{figure}

The Spaun 2.0 model was deployed on the NVIDIA GPU system. To achieve a real-time multi-GPU implementation, the original Spaun 2.0 architecture was adapted to partition each of Spaun’s modules into its own ``sub-network''. Each of these sub-networks can then be run in its own NengoOCL simulator \citep{nengo}. Figure \ref{fig_spaun_gpu} illustrates how the Spaun 2.0 system architecture has been distributed across 7 GPUs. 
In the future, this system could be deployed on liquid-cooled neuromorphic hardware, such as SpiNNaker2, in order to make a full use of its energy-efficiency and further improve the latency.
While the original Spaun 2.0 architecture contains 8 modules (which could intuitively be distributed across 8 GPUs), the reward evaluation and action selection modules have been combined due to the lightweight structure of the reward evaluation system and its extensive connectivity with the action selection system.

\subsubsection{Conversational engine}

Seamless low-latency conversations with the social robot necessitate a high-performance large language model.
To this end, a recent model from the LLaMA line introduced by Meta was used \citep{touvron2023llama2}.
For the purpose of this showcase we used the variant of the LLaMA model with 13B parameters, which allows for quick computation of the responses without sacrificing their quality. This model is deployed locally on a GPU system.

The robot can have a multi-turn conversation, answer general domain questions, and questions related to the showcase within the context information provided to the model.
In order to enable the robot to communicate with users from different regions we use translation software. Verbal communication with the robot is translated using Google Cloud Translation API.

\subsubsection{Speech generation}

In order to allow the text produced by the language models to be converted into lifelike speech, and thus enable the robot to communicate with its users, we decided to use Amazon Polly API \citep{Polly_Amazon}.
Amazon Polly employs advanced deep learning technologies to generate human-like speech, enabling the conversion of written text into natural-sounding audio. 

\subsubsection{Speech recognition}

The OpenAI Whisper model \citep{Whisper} is used to enable the robot to listen to commands spoken by the users.
Whisper is an automatic speech recognition system which shows human level robustness and accuracy on English speech recognition. 
The Whisper model has 1550M parameters and runs locally on one of the NVIDIA DGX A100 machines.

\subsubsection{Player detection and depth analysis}
\label{cameras}

To enhance the robot's awareness of its environment we use three strategically placed cameras around the stage. Each camera employs a distinct method to interpret the scene:

\begin{itemize}
\item Camera 1 focuses on the audience in front of the stage, utilizing real-time face detection to count the number of people looking at the stage.

\item Camera 2 is tasked with detecting the presence of people on the stage. This can be used to understand stage occupancy and ensure safety measures are followed during Ameca's performance.

\item Camera 3 incorporates a depth sensor to identify if someone is located behind the theremin. This gives the robot an indication to interact with the theremin player. To this end, we used a state-of-the-art depth estimation model, Depth Anything, introduced by \cite{depthanything}.
\end{itemize}

\subsubsection{Intention detection}

Intention detection to drive the robot's FSM and enable interactive musical experience was carried out using the GPT3.5 API provided by OpenAI \citep{GPT3}. 
Intention detection works by specifying a context and one or more intentions derived from it. The possible intentions consist of a name and a description combined as key-value pairs in a dictionary. These records are passed to an LLM together with the context and the message from which to detect the intention. A callback function handles the output which is the name of the detected intention or “None” if no intention could be detected.

\subsubsection{Hand tracking}

To visualise the user's input during the interactive theremin duet with the Ameca robot we decided to develop energy-efficient, low-latency hand tracking.
Tracking human hands is key for many robotic applications like imitation learning, human-robot collaboration, and it plays a key role in the showcase of playing an instrument with a human.  
To this end, we collaborated with Intel and employed their latest neuromorphic processor - Loihi2 \citep{davies_intel} in conjunction with a DVS camera.
Given that an event camera only outputs information when a change in scene brightness is detected at a particular pixel, a fixed event camera can readily serve as an elementary motion detector. As a result of motion between the scene and the camera, the pixels independently report perceived changes in brightness, resulting in a sparse and asynchronous stream of spatiotemporal data. See Section \ref{realtime_proc} for more details regarding the hardware setup.
For the purpose of this project, three parallel tracks were developed to enable reliable hand tracking. 

\begin{figure}[H]
\centering
\includegraphics[width=0.9\textwidth]{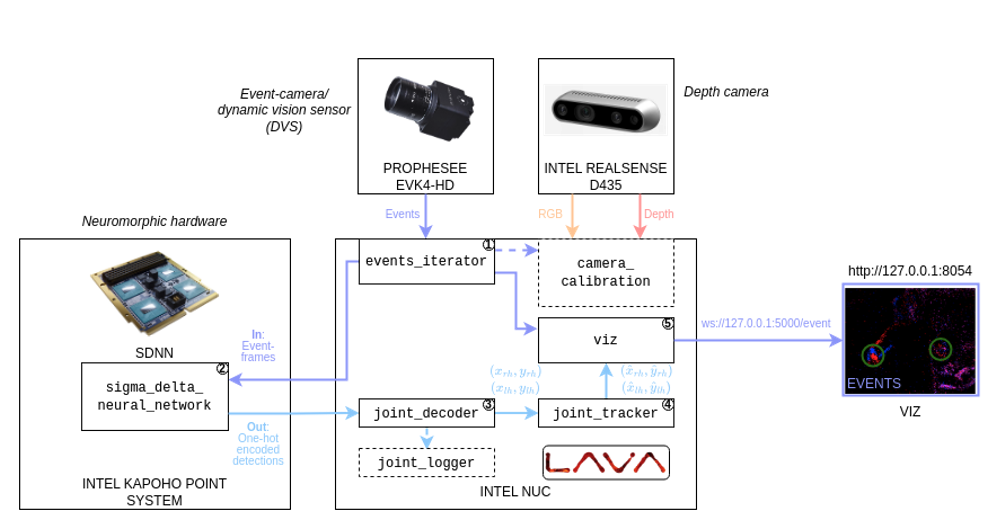}
\caption{Overview of the software layer of the neuromorphic hand tracking.}
\label{fig_loihi_software}
\end{figure}

The first track deploys a deep spiking neural network (DSNN) trained to track hands using the DHP19 event-based dataset which features automatically generated labels for joints of a moving person that were obtained with a motion tracking system \citep{dhp19}. We used the dataset to train a spiking neural network compatible with Loihi2 to track the hands. 
By using spikes as the basis for computation, neuromorphic hardware achieves the highest efficiencies when running appropriate, spike-based algorithms. Given that the state-of-the-art neural-network-based motion tracking has been optimized for a different type of hardware (e.g., GPUs), novel neural network architectures are needed to take full advantage of neuromorphic computing platforms. The sigma-delta quantization is an example of an effective approach to run convolutional neural network more efficiently on conventional, as well as neuromorphic, hardware. In sigma-delta neural networks, spiking is achieved by means of a delta encoding process and a sigma decoding process. The encoder computes the difference between the latest nonlinear activations of the ANN neurons. Once the difference exceeds a predefined threshold, the encoder sends a spike to the next layer. The decoder accumulates the spikes until the original value is recovered, providing a discrete approximation of the ANN activation process.

\begin{figure}[H]
\centering
\includegraphics[width=0.8\textwidth]{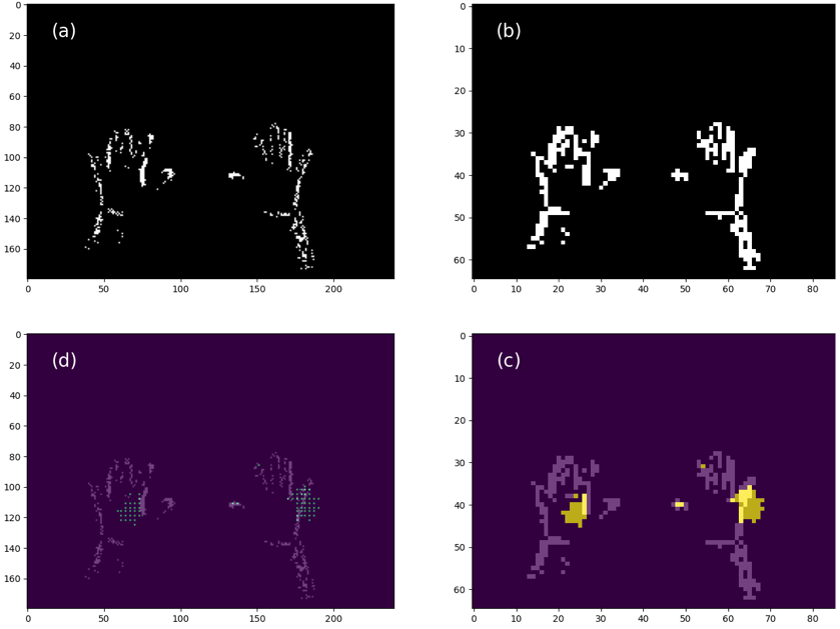}
\caption{Tracking example with both hands moving side-to-side (waving): (a) (240x180) event-frame capturing the active hands; (b) (86x65) event-frame downsampled to run on chip; (c) output of the multi-peak DNF overlayed the downsampled event-frame; (d) upscaled DNF output overlayed the original event-frame.}
\label{fig_loihi_DNF}
\end{figure}

The second computation that is deployed relies on the theory of dynamic neural fields \citep{dynamic_field} to improve deep-learning based tracking. This neuromorphic algorithm can also be used to filter the coherent motion in the event-stream, constituting an additional mechanism to generate hand-motion hypotheses. Dynamic field theory (DFT) is a mathematical and conceptual framework for modelling human cognition. This approach has been used to develop cognitive architectures for robots (spatial language, action planning, learning), and is used as one of computing paradigms for high-level programming of neuromorphic chips. 
In this framework, neural populations (layers) form attractor dynamics with stabilized spatial patterns that filter out noise and provide temporal filtering for tracking moving objects in the presence of distractors.

In this application, we decided to use these methods in a sequential configuration, in which the DNF-filter stabilizes the output of the deep neural network. Additional filtering of the visual scene (figure-background segmentation) is performed based on the depth information from a conventional RGB-D camera, in this case Intel’s RealSense (see fig. \ref{fig_loihi_DNF}).


\section*{Author contributions}
Authors are grouped based on contributions:
T.R., X.C., C.M., H.G., M.A., K.K., C.E., M.A. - Initial programming, theremin research, and Spaun integration. 
M.H., S.F. - Help with writing the paper.
Y.S., L.A., S.G. - Development of the Loihi2 subsystem.
J.F., L.F., F.W. - Programming and further development of the showcase. 
L.W., H.P., J.S., K.R., N.A., M.H. - Supervision and original vision for the showcase. 

\section*{Acknowledgements}

We would like to also thank the NEOM smart city initiative for initiating and funding the creation of the Theremin playing robotic usecase. We thank Intel Neuromorphic Research Community and Intel's Neuromorphic Computing Lab for their support with Loihi2 implementation. Some of the ideas behind this work were originally discussed by the authors at the Capo Caccia Workshop towards  Neuromorphic Intelligence. \\

Moreover,  we would like to thank and acknowledge the role of our colleagues and collaborators:
Prof. Dr. Chris Eliasmith,
Carolina Eyck,
Nourchene Ferchichi,
Matthieu  Amiguet,
Travis Dewolf,
Eric Schulz, 
Eric Wallin,
Mike Davies,
and Morgan Roe.

\bibliography{biblio.bib} 

\clearpage

\end{document}